\documentclass[conference]{IEEEtran}
\IEEEoverridecommandlockouts
\usepackage{cite}
\usepackage{amsmath,amssymb,amsfonts}
\usepackage{algorithmic}
\usepackage{graphicx}
\usepackage{textcomp}
\usepackage{xcolor}
\usepackage{algorithm}

\usepackage{booktabs}
\usepackage{tablefootnote}
\usepackage{bbding}
\usepackage{color}
\usepackage{float}
\usepackage{color}
\usepackage{makecell}
\usepackage{multirow}
\usepackage{hyperref}

\def\BibTeX{{\rm B\kern-.05em{\sc i\kern-.025em b}\kern-.08em
    T\kern-.1667em\lower.7ex\hbox{E}\kern-.125emX}}
\begin{document}

\title{BEAR: A Video Dataset For Fine-grained \\ Behaviors Recognition Oriented \\ with Action and Environment Factors}

\author{\IEEEauthorblockN{Chengyang Hu}
\IEEEauthorblockA{\textit{Shanghai Jiao Tong University} \\
Shanghai, China}
\and
\IEEEauthorblockN{Yuduo Chen}
\IEEEauthorblockA{\textit{Shanghai Jiao Tong University} \\
Shanghai, China}
\and
\IEEEauthorblockN{Lizhuang Ma$^{\ast}$ \thanks{* Corresponding author. E-mail: ma-lz@cs.sjtu.edu.cn.}}
\IEEEauthorblockA{\textit{Shanghai Jiao Tong University} \\
Shanghai, China}}

\maketitle

\begin{abstract}
Behavior recognition is an important task in video representation learning. An essential aspect pertains to effective feature learning conducive to behavior recognition. Recently, researchers have started to study fine-grained behavior recognition, which provides similar behaviors and encourages the model to concern with more details of behaviors with effective features for distinction. However, previous fine-grained behaviors limited themselves to controlling partial information to be similar, leading to an unfair and not comprehensive evaluation of existing works. In this work, we develop a new video fine-grained behavior dataset, named BEAR, which provides fine-grained (\textit{i.e.} similar) behaviors that uniquely focus on two primary factors defining behavior: Environment and Action.  It includes two fine-grained behavior protocols including Fine-grained Behavior with Similar Environments and Fine-grained Behavior with Similar Actions as well as multiple sub-protocols as different scenarios. Furthermore, with this new dataset, we conduct multiple experiments with different behavior recognition models. 
% Our research focuses on a widely used trick, input modality, which influences the study of these two essential factors, environment and action, in behavior recognition. 
Our research primarily explores the impact of input modality, a critical element in studying the environmental and action-based aspects of behavior recognition.
Our experimental results yield intriguing insights that have substantial implications for further research endeavors.
\end{abstract}


\section{Introduction}
\begin{quote}
\textit{``Behavior is range of \textbf{actions} made by individuals in some \textbf{environment}."}\cite{quote1}
\end{quote}

% Recently, behavior (\textit{i.e.} action) recognition has been a widely researched topic in video representation learning. 
The noteworthy advancements in behavior (\textit{i.e.} action) recognition as a widely researched topic went viral in the domain of video representation learning. To explore different recognition models for behavior understanding, different benchmarks (\textit{i.e.} datasets) ~\cite{activitynet,kinetics} were released as consistent standards for evaluating behavior recognition technologies. 

\begin{figure}[t!]
    \centering
    \includegraphics[width=0.48\textwidth]{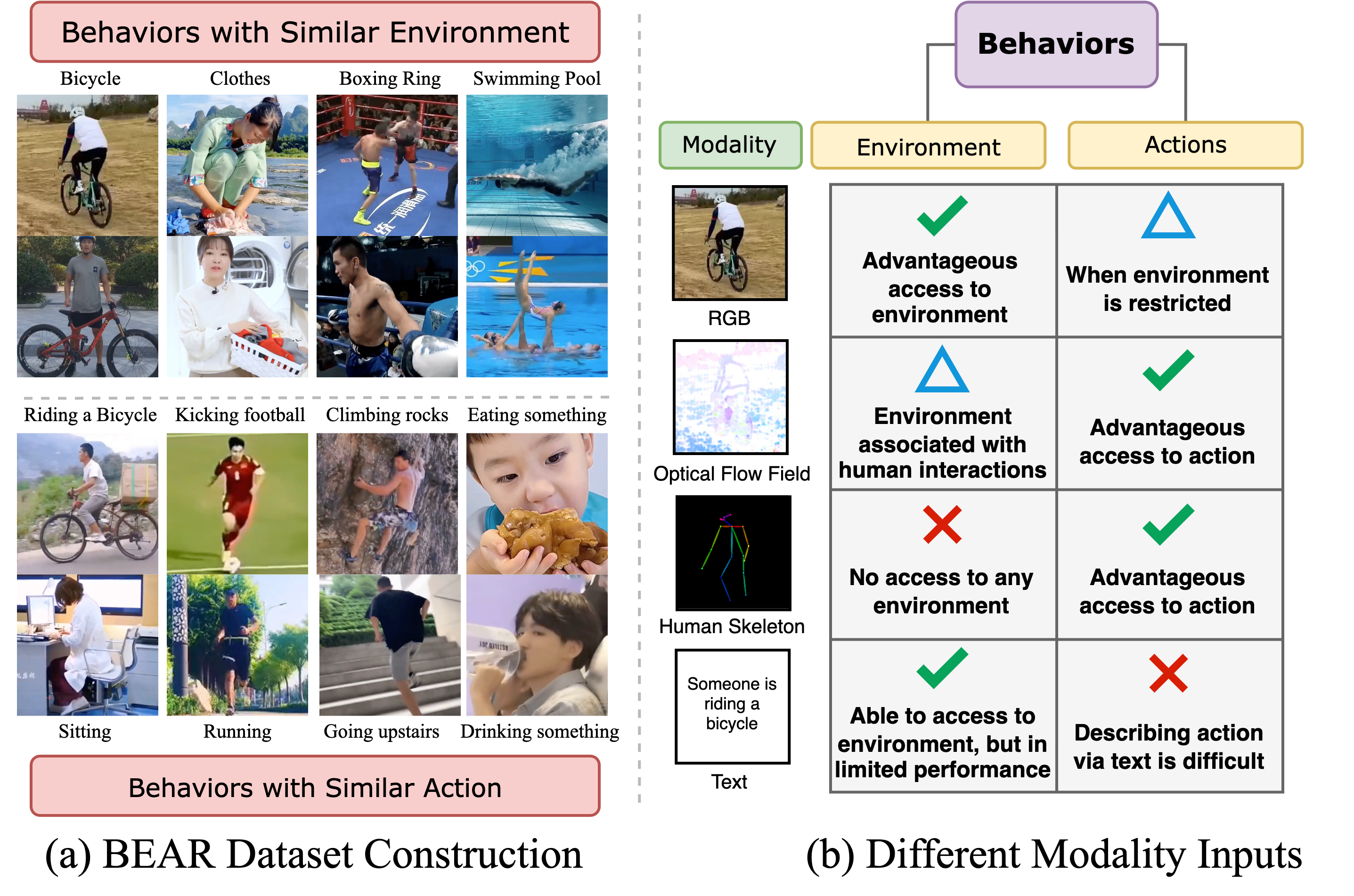}
    \caption{BEAR dataset for exploring the effectiveness of different modalities in two important factors of behavior recognition -- environment and action. (a) The protocols of BEAR are in well-controlled conditions with the environment and action. (b) The conclusion of the information is that different modalities are learned for behavior recognition. \textcolor{black}{\Checkmark}: Modality can learn this factor. \textcolor{black}{\XSolidBrush}: Modality cannot learn this factor. \textcolor{black}{$\triangle$}: Modality can learn this factor in some condition.}
    \label{fig:fig1}
    %\vspace{-5mm}
\end{figure}

Researchers initially focus on general behavior recognition, such as HMDB51~\cite{hmdb}, UCF101~\cite{ucf101}, Kinetics~\cite{kinetics} and ActivityNet~\cite{activitynet}, that only roughly consider the behaviors. Although many methods~\cite{tsm,tsn,videomae} show promising results on these benchmarks, they easily categorize different behaviors based on partial information, \textit{e.g.} background information, which leads to the exploration of fine-grained behavior recognition that should extract essential features of the behaviors. Fine-grained behavior recognition aims to provide more specific behaviors that are similar to distinguish and challenge the recognition technology to focus on the detailed information of these behaviors. Previous works have provided multiple benchmarks like Breakfast~\cite{breakfast}, MPII-Cooking 2~\cite{mpii}, Diving 48~\cite{diving48}, Something-something V2~\cite{ssv2} and FineGym~\cite{finegym}. Nevertheless, these benchmarks arbitrarily assume some factors in behaviors are detrimental to behavior recognition. Some works~\cite{breakfast,mpii,finegym} release data in similar environments to eliminate environment's influence. For instance, FineGym collects different actions in a gym environment (\textit{e.g.} ``Uneven Bars" and ``Balance Bean") and sub-actions (\textit{e.g.} "Beam Turns" and ``Leap-Jump-Hop" in ``Balance Bean"). Also, some work~\cite{ssv2} tries to puzzle the model with similar actions but interact with different environments. Something-something V2 provides similar actions that interact with different ``something", like ``Putting Something on a Surface".

To explore fine-grained behavior recognition fairly, inspired by Behavioral Economics studies~\cite{quote1}, we first concern two important factors that construct a behavior: 1) \textbf{Actions.} Actions reflect how the human physical systems (like the body, and limbs) change objectively. 2) \textbf{Environments.} Environments show the objects that individuals interact with and the place where the individual stays. In this study, we construct a new fine-grained behavior dataset named \textit{\textbf{B}ehaviors for \textbf{E}nvironment and \textbf{A}ctions \textbf{R}ecognition dataset} (denoted as \textbf{BEAR}\footnote{Project page: https://hu-cheng-yang.github.io/projects/ICME25\_BEAR/}), which is a well-controlled dataset that provides multiple paired behaviors with similar actions or similar environments to confuse current classification models.

With proposed dataset, we execute a series of experiments involving distinct behavior recognition models. For video behavior recognition models, one widely used trick to improve the model generalization is utilizing different input modality (shown in Figure~\ref{fig:fig1} (a)), \textit{e.g.} RGB, optical flow field~\cite{opt1,opt2}, human skeleton~\cite{pose1,pose2,pose3}. With the development of the vision-text foundation model~\cite{clip,videoclip}, the text is also a useful modality in behavior recognition. One example that explores the effectiveness of modality is TSN~\cite{tsn}, which disentangles behaviors into spatial and temporal to extract the features by both RGB input and optical flow fields. In this work, we explore the effectiveness of different modality inputs from a new standpoint -- \textit{What kind of information do different modalities provide for fine-grained behavior recognition, environment or action information}? Our observations (shown in Figure~\ref{fig:fig1} (b)) is listed in Section~\ref{summary}.
% include: 1) RGB learns action information until the environment is restricted, or RGB focuses more on environment information. 2) The optical flow field learns the action and environment associated with human interactions, which is a robust modality. 3) Solely relying on the human skeletal modality results in the acquisition of purely action information. 4) Text modality demonstrates a propensity toward environmental information. However, it remains a substantial distance away from achieving proficiency in behavioral recognition.

Our work contributes to the following aspects:
\begin{itemize}
    \item We develop a fine-grained dataset BEAR with controlled environment and action factors which provides comprehensive protocols to explore information that different modalities learn for behavior recognition.
    \item We conduct extensive experiments on BEAR and analyze how the existing modality information contributes the robust feature learning, which may provide new principles for future research in behavior recognition.
\end{itemize}

\begin{figure*}
    \centering
    \includegraphics[width=\textwidth]{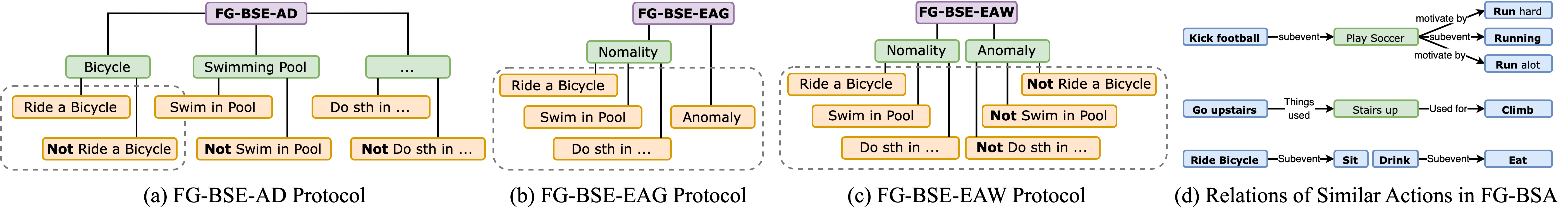}
    \caption{Protocols of FG-BSE and FG-BSA. (a) FG-BSE-AD Protocol. (b) FG-BSE-EAG protocol. (c) FG-BSE-EAW protocol. The dotted block shows the categories used in this setting. (d) The relations of similar actions from ConceptNet. The arrow $\rightarrow$ indicates the relations between different behaviors. All pairs in the FG-BSE-BSA setting have similar actions semantically.}
    \label{fig:protocol}
\end{figure*}

\section{Related Work}

\noindent\textbf{Datasets in Behavior Recognition.} As the foundation of video representation learning, more and more high-quality datasets come out. First, datasets like HDMB51~\cite{hmdb}, UCF101~\cite{ucf101}, and ActivityNet~\cite{activitynet} provide lots of videos in different categories.
% and consider video representation learning as a classification task. 
% However, these datasets are coarse-grained which may extract un.
% To mitigate the effectiveness of environment information
To explore how to extract more detailed and effective features, some work tries to form fine-grained datasets.
However, these benchmarks only restrict partial information to form similar behaviors. Benchmarks like Breakfast~\cite{breakfast}, MPII-Cooking2~\cite{mpii}, Diving 48~\cite{diving48}, and FineGym~\cite{finegym}, provide the behaviors that contains the similar environments. Also, benchmarks like Something-something V2~\cite{ssv2} form similar behaviors with similar actions but interact with different environment objects.
% like Breakfast~\cite{breakfast}, MPII-Cooking2~\cite{mpii}, Diving 48~\cite{diving48}, and FineGym~\cite{finegym}. 
% However, it is naive to consider that environment is redundant. Even if the action is similar, it is different behavior because they interact with different objects or occur in different places. 
Considering these issues, our work builds a new fine-grained dataset named BEAR with a well-controlled environment and action factors and provides a detailed analysis of how the different modality information extracts these two factors in behavior recognition.
% The detailed comparison of existing datasets is shown in Table~\ref{tab:datasets}.

% \noindent\textbf{Behavior Recognition.} Initially, previous methods~\cite{2dcnn1} adopt 2D CNNs, thereby reformulating the behavior recognition into image classification task. 
% However, the efficacy of these methods is compromised due to the disregard of temporal information. Addressing this limitation, some researchers harness more potent backbones, like 3D CNNs~\cite{3dcnn1,3dcnn2}, RNN~\cite{rnn1,rnn2}, and Transformer~\cite{transformer2,videomae}. 
% While yielding more impactful outcomes, these methods lack interpretability.
% Another way is utilizing different modality inputs.
% Beyond the RGB, some methods provide modalities like optical flow fields~\cite{opt1,tsn}, human skeleton~\cite{pose2,pose3}, and even text~\cite{videoclip}. 
% While these approaches with various modality inputs offer promising results, they fall short of providing a comprehensive analysis of how these appended modalities enhance the extraction of robust features.

\noindent\textbf{Modality Inputs in Behavior Recognition.} In early research, RGB~\cite{3dcnn1,3dcnn2} is the most common input for ease of obtaining. However, to improve the effectiveness of feature learning, different modality inputs start to emerge. The most two popular modalities are the optical flow field and the human skeleton. The optical flow field describes the pattern of apparent motion between two frames~\cite{opt1,tsn}. The human skeleton shows the key points of the human part in the images~\cite{pose1,pose2}. 
These modalities improve performance for behavior recognition. Recently, with the development of the text-vision foundation model, some works~\cite{videoclip} utilize text as a modality to guide the model learning video presentation. In this work, we will analyze how these modalities contribute to learning the environment and action factors.

\section{The BEAR Dataset}

% The BEAR dataset aims to analyze how the multi-modality extracts effective features to learn decisive factors, environment and actions, in behavior recognition. For this reason, we collect the data with similar environments and actions with comprehensive protocols to evaluate the performance of recognition models with different modalities inputs. 
The primary objective of the BEAR dataset is to 
% explore how different modality inputs prompt impactful features for comprehending critical factors such as environments and actions in fine-grained behavior recognition. 
release a more fair and comprehensive fine-grained video behavior dataset by controlling the two decisive factors of behavior: environment and action.
This endeavor seeks to offer a rigorous benchmark accompanied by meticulously crafted annotations, contributing to the sphere of behavior recognition.
Compared to previous works~\cite{hmdb,ucf101,breakfast,finegym}, BEAR has the following features: 1) \textbf{Comprehensive protocols with controlled factors.} We provide multiple protocols to evaluate the performance of modalities to recognize the behaviors in similar environments and similar actions. 2) \textbf{Multiple environments and actions with different behaviors.} We provide 8 similar environments and 4 pairs of similar actions in different behaviors to evaluate the models in a fine-grained way and explore how the environment and actions will be concerned by different modality inputs. 3) \textbf{Wild data with rich variations.} Even with the same behavior, we collect data with different quality, different scales, and different shooting perspectives, in a real-world setting. These features distinguished our dataset from vanilla action recognition datasets, which are more challenging for behavior recognition.

\subsection{Dataset Protocols}

In BEAR dataset, we provide two fine-grained behavior recognition protocols: Fine-grained Behaviors with Similar Environments (FG-BSE) and Fine-grained Behaviors with Similar Actions (FG-BSA). Also, we provide different sub-protocols as more difficult scenarios. For all protocols, we keep $\sim 70\%$ of the video data as the training dataset, and the left $\sim 30\%$ as the testing dataset. To avoid model classifying the behaviors via the video ID, the data in the training and testing dataset are from the different videos.

\noindent\textbf{Fine-grained Behaviors with Similar Environments.} In FG-BSE protocols, we provide three sub-protocols: \textbf{1)} Anomaly Detection (FG-BSE-AD) protocol. In the FG-BSE-AS protocol, we provide the pairs of normality and anomaly behaviors with similar environments and ask the model to tell these two behaviors, which can be considered as a binary classification task. In this setting, we provide eight environments with different interact objects (denoted as obj.) 
% like \textit{``Bicycle", ``Clothes", ``Football", ``Phone", ``String instruments"} 
or places (denoted as pl.).
% like \textit{``Boxing ring", ``Mountain rock", ``Pool"}. 
For instance,  we define the normality behavior of ``Bicycle" as riding a bicycle and the anomaly behavior as interacting with a bicycle but not riding. We conduct eight experiments with the previous setting. The detailed definition of normality behavior is shown in Figure~\ref{fig:protocol}~(a) and Appendix. The evaluation metrics are Equal Error Rate (EER) and Area Under the Curve (AUC). \textbf{2)} Environment AGnostic (FG-BSE-EAG) protocol. In the FG-BSE-EAG protocol, as shown in Figure~\ref{fig:protocol}~(b), we form a multi-class classification that introduces all normality and anomaly behaviors from previous settings. Normality behaviors are considered as different behaviors, and anomaly behaviors are considered as only one category no matter this sample from any environment. Following this definition, there are 9 categories, with eight for normality and only one for anomaly behavior. The evaluation metrics are Top-1 and Mean-1 accuracy. \textbf{3)} Environment AWare (FG-BSE-EAW) protocol. In FG-BSE-EAW protocol, the anomaly behaviors from different environments are also considered as individual categories, as shown in Figure~\ref{fig:protocol}(c). The evaluation metrics are Top-1 and Mean-1 accuracy.

\noindent\textbf{Fine-grained Behaviors with Similar Actions.} In FG-BSA protocols, one important issue is to determine the definition of ``similar action". We utilize the semantic definition of different behaviors and composite the pairs with similar semantic meanings in action. To deal with this issue, we introduce ConceptNet~\cite{speer2017conceptnet} and find the behavior pairs with semantic relations (\textit{e.g.} subevent) in action, which indicates the pairs have similar actions and can be transformed in different environments. With these rules, we define four pairs of fine-grained behavior pairs with similar actions, including \textit{``Running / Kicking football", ``Climbing rocks / Going upstairs", ``Riding a bicycle / Sitting", ``Eating something / Drinking something"}. The relation of these behavior pairs in ConceptNet is shown in Figure~\ref{fig:protocol}(d). FG-BSA protocol provides one protocol: Binary Classification (FG-BSA-BC) protocol. Similar to FG-BSA-AD protocol, FG-BSA-BC aims to evaluate the models to distinguish pairs of behaviors as a binary classification task. The evaluation metrics are EER and AUC. 

\begin{table*}[t]
\caption{Comparison result on FG-BSE-AD protocol. We compare the model with different modality inputs, like RGB: RGB input, Flow: Optical flow field, Skeleton: Human skeleton. ``obj" indicates the environment focus on the interacting object, ``pl" indicates the environment focus on occurring place. \textbf{Bolded} values shows the best result, \underline{underlined} ones are the second best.}\label{setting1_1}
\centering
\scalebox{0.97}{
\begin{tabular}{@{}l|c|cccccccc|cc@{}}
\toprule
\multirow{2}{*}{\textbf{Method}} & \multirow{2}{*}{\textbf{Modality}} & \multicolumn{2}{c}{\textbf{Bicycle}~[obj]}       & \multicolumn{2}{c}{\textbf{Clothes}~[obj]}       & \multicolumn{2}{c}{\textbf{Boxing ring}~[pl]}       & \multicolumn{2}{c}{\textbf{Swimming pool}~[pl]} & \multicolumn{2}{|c}{\textbf{Average}}       \\ \cmidrule(l){3-12} 
                                 &                                  &                                    \textbf{EER(\%)} & \textbf{AUC(\%)} & \textbf{EER(\%)} & \textbf{AUC(\%)} & \textbf{EER(\%)} & \textbf{AUC(\%)} & \textbf{EER(\%)} & \textbf{AUC(\%)} & \textbf{EER(\%)} & \textbf{AUC(\%)} \\ \midrule
C3D~\cite{C3D}                              & RGB                                                             &         22.66         &        87.73          &      34.38            &        66.04          &        \underline{23.19}          &      81.57            &          17.95        &         92.42    & 25.00 &  82.06     \\
SlowFast~\cite{slowfast}                         & RGB                                                        &      33.10            &        63.95          &     41.89             &        60.79          &        45.22          &        58.90          &            25.00      &         82.24    & 36.30 &  66.47     \\
TSM~\cite{tsm}             & RGB                                                  &                  37.50&60.79&31.24&78.33&35.39&72.14&35.80&71.90    & 34.98 &     70.79          \\
VideoMAE~\cite{videomae}                         & RGB                                                       &         45.49         &      53.32            &      43.11            &         59.94         &         35.12         &        65.02          &           24.48       &          82.62    &   37.05   &   65.23   \\ \midrule
\multirow{3}{*}{TSN~\cite{tsn}}             & RGB                                                 &                  \underline{19.78}&\underline{87.03}&23.73&81.33&24.61&78.20&\underline{14.11}&\textbf{93.63}         & \underline{20.56} &   85.05         \\
                                 & Flow                                                       &        22.93          &       85.95          &       22.51           &     83.02             &    \textbf{17.88}              &     \underline{89.49}             &        27.18          &     81.68  & 22.63 &      85.04       \\
                                 & RGB+Flow                         &       \textbf{15.49}                 &      \textbf{93.18}            &       \textbf{18.34}           &   \textbf{90.62}               &       28.36           &     \textbf{92.76}             &                      15.81            &       91.05   & \textbf{19.50} &   \textbf{91.90}     \\ \midrule
PoseC3D~\cite{posec3d}        & Skeleton                                          &         20.05       &       82.11           &                22.51&86.49&28.50&84.53&\textbf{13.66}&\underline{92.99}      & 21.18 &      \underline{86.53}        \\
STGCN~\cite{STGCN}                            & Skeleton                                               &          31.76          &      70.73     &       \underline{19.37}&\underline{84.15}&29.27&82.14&16.38&90.45    & 24.20 &       81.87       \\ \bottomrule
\end{tabular}}

\end{table*}

\section{Empirical Studies}

With the BEAR dataset, 
% we conduct experiments to explore how the multi-modality input influences the environment and action feature learning in behavior recognition 
 we undertake empirical investigations to elucidate the impact of different modualities inputs on the learning of environment and action information in the domain of behavior recognition.
All experiments are employed via MMAction~\cite{mmaction2}. 
% We use the default training setting for these methods and the details can be found in Appendix.
% Also, we provide multiple analyses and observations as a direction on how to utilize multi-modality information for further research.
% Furthermore, our study encompasses diverse analyses and observations, serving as a guidepost for harnessing different modality inputs in the pursuit of advanced research.
% The experimental setting can be found in Appendix.
We conduct experiments on well-known methods with RGB input models, like C3D~\cite{C3D}, SlowFast~\cite{slowfast}, TSM~\cite{tsm}, and VideoMAE~\cite{videomae}, models with optical flow field as input, like TSN~\cite{tsn}. We also conduct experiments with multi-modality input, RGB and optical flow field on TSN. And models with human skeleton as input like PoseC3D~\cite{posec3d} and STGCN~\cite{STGCN}. Also, to show the effectiveness of the text input, we also conduct zero-shot testing on VideoCLIP~\cite{videoclip}. The detailed implementations are shown in Appendix.

\begin{table}[t]
\caption{Comparison result on FG-BSE-EAS and FG-BSE-EAW protocols. \textbf{Bolded} values shows the best result, \underline{underlined} ones are the second best result.}\label{setting1213}
\centering
\scalebox{0.98}{
\begin{tabular}{@{}l|cc|cc@{}}
\toprule
\multirow{2}{*}{\textbf{Method}} & \multicolumn{2}{c}{\textbf{FG-BSE-EAS}} & \multicolumn{2}{|c}{\textbf{FG-BSE-EAW}} \\ \cmidrule(l){2-5} 
                                 & \textbf{Mean-1(\%)}      & \textbf{Top-1(\%)}      & \textbf{Mean-1(\%)}     & \textbf{Top-1(\%)}    \\ \midrule
C3D~\cite{C3D}                              &              \textbf{74.02}           &      \textbf{73.06}                   &            \textbf{59.90}            &         \textbf{74.33}            \\
SlowFast~\cite{slowfast}                         &             62.04            &        57.96                &            36.68           &          56.11             \\
TSM~\cite{tsm}             &          \underline{73.75}              &            \underline{72.84}       &               \underline{61.16}         &            \underline{72.42}           \\
VideoMAE~\cite{videomae}                         &       43.34                &           43.76           &                 29.83       &             49.17          \\ \midrule
TSN(RGB)~\cite{tsn}             &              72.42           &     72.29                  &            56.21            &          71.59            \\
TSN(Flow)~\cite{tsn}             &              53.31           &     61.66                   &            38.70            &           58.79            \\
TSN(RGB+F)~\cite{tsn}             &          62.60 &	71.53 &	55.47 &	72.99            \\

\midrule
PoseC3D~\cite{posec3d}         &             46.60           &           51.70              &           30.24            &           51.61            \\
STGCN~\cite{STGCN}              &   59.90         &          62.32            &                   37.88           &           52.41                                  \\ \bottomrule
\end{tabular}
}

\end{table}

\subsection{Comparison Result of FG-BSE Protocol}
% \subsubsection{Comparison Result of FG-BSE Protocol}

\noindent\textbf{Result in FG-BSE-AD Protocol.}
As a result of FG-BSE-AD protocol shown in Table~\ref{setting1_1}, we have the following observations: 1) Generally, methods with optical flow field and human skeleton inputs perform better. It indicates that the optical flow field and human skeleton inputs contain more action-related information which is an essential issue for fine-grained behaviors with similar environments. 2) For RGB inputs, pre-train weights play an important role. For C3D and TSN, which perform better than other RGN input methods, they use Sports1M~\cite{sports1m} and ImageNet~\cite{deng2009imagenet} as default pre-train weights. Another observation is that VideoMAE with a powerful backbone does not show promising performance. The reason is that the ViT backbone needs plenty of data to learn the effective features, which is not suitable for our data's quantity. 3) Multi-modality input (TSN) shows a promising result and leads to the state-of-the-art on average. Although the optical flow field and human skeleton show a promising result, collaborating with RGB, they can extract more features for better performance. One explanation is that RGB contains undetermined features for classification, however, collaborating with the optical flow field, can provide more information when it is hard for decisions with only the optical flow field.

% When comes to environment-agnostic and environment-aware settings, things are going differently. We have the following observations
\noindent\textbf{Result in FG-BSE-EAS and FG-BSE-EAW Protocol.} In the context of FG-BSE-EAS and FG-BSE-EAW Protocols, as shown in Table~\ref{setting1213}. Compared to FG-BSE-AD, divergent observation emerges, giving rise to the subsequent founds:
1) The most significant observation is that the methods with optical flow field and human skeleton input underperform the methods with RGB inputs. One reason is that the optical flow field and human skeletons have less environment information, which failed to classify these behaviors with a significant environment divergence.
% 2) Generally, the performance on FD-BSE-EAS shows a better performance than the FD-BSE-EAW, indicative that the latter scenario is more challenging. For the anomaly behaviors are more inconsistent than normality ones, which is harder for classify when consider them as individual categories.
% which indicates the latter setting is more challenging. 
2) Even for the best performance in C3D, approximately 70\% of the data can be accurately categorized, yielding a less promising outcome.
3) Multi-modality inputs fail in FG-BSE-EAS and FG-BSE-EAW Protocols. TSN with both RGB and the optical flow field inputs only reflects a similar performance with TSN with RGB inputs. It shows that when the environment is more diverse, the minor differences between the actions are not sufficient to classify. 
% However, the result with multi-modality is still acceptable. From all protocols in FG-BSE, one conclusion is that multi-modality is necessary for behavior recognition because different modalities will make different contributions in various scenarios. 
% only about 70\% of data can be correctly classified, which is not a promising result.

\begin{table*}[t]
\caption{Comparison result on FG-BSA-BC protocol. We compare the model with different modality inputs. RGB: RGB input, Flow: Optical flow field, Skeleton: Human skeleton. Run/Kick: Running / Kicking football, Climb/Stair: Climbing rocks / Going stairs, Bicycle/Sit: Riding a bicycle / Sitting, Eat/Drink: Eating something / Drinking something. \textbf{Bolded} values shows the best result, \underline{underlined} values shows the second best result.}\label{setting21}
\centering
\scalebox{1.0}{
\begin{tabular}{@{}l|c|cccccccc|cc@{}}
\toprule
\multirow{2}{*}{\textbf{Method}} & \multirow{2}{*}{\textbf{Modality}} & \multicolumn{2}{c}{\textbf{Run/Kick}}       & \multicolumn{2}{c}{\textbf{Climb/Stair}}       & \multicolumn{2}{c}{\textbf{Bicycle/Sit}}       & \multicolumn{2}{c}{\textbf{Eat/Drink}}   & \multicolumn{2}{|c}{\textbf{Average}}    \\ \cmidrule(l){3-12} 
                                 &                                  &                                \textbf{EER(\%)} & \textbf{AUC(\%)} & \textbf{EER(\%)} & \textbf{AUC(\%)} & \textbf{EER(\%)} & \textbf{AUC(\%)} & \textbf{EER(\%)} & \textbf{AUC(\%)} & \textbf{EER(\%)} & \textbf{AUC(\%)} \\ \midrule
C3D~\cite{C3D}                              & RGB                                                           &          \textbf{5.90}        &          \underline{98.69}        &      7.65            &         98.12         &          8.49        &        96.92         &       12.58           &         94.94     &  8.66  &  97.17  \\
SlowFast~\cite{slowfast}                         & RGB                                                        &          9.75        &        94.31          &       31.86           &         76.33         &           14.46       &         94.31         &        17.09         &        91.03     & 18.29 &  89.00  \\
TSM~\cite{tsm}             & RGB                                                                &      10.26            &           92.94       &        28.76          &         78.56         &            12.66      &         92.48         &   16.55  &       92.44   & 17.06 &  89.11 \\
VideoMAE~\cite{videomae}                         & RGB                                                       &         11.01         &         92.90         &       41.40           &         61.19         &         16.40         &         91.96         &         16.56          &        88.75      & 21.34 &  83.70  \\ \midrule
\multirow{3}{*}{TSN~\cite{tsn}}             & RGB                                                       &        9.75          &     96.34             &         \underline{5.10}         &        \underline{98.89}          &      \underline{4.36}            &        \underline{99.29}          &         \textbf{6.85}         &      \underline{97.25}      &\underline{6.52} &    \underline{97.63}  \\
                                 & Flow                                                     &  8.72                &     96.82           &    12.74            &  94.44                  &    14.46              &       91.27             &         14.36         &         93.39         & 12.57 & 93.98\\
 & RGB+Flow                                         & \underline{6.93} &	\textbf{98.75}            &   \textbf{4.83} 	&  \textbf{99.47}                 &   \textbf{2.54} &	\textbf{99.53}            &     \underline{10.40} &	\textbf{97.66}        & \textbf{6.18} &	\textbf{98.85} \\\midrule
PoseC3D~\cite{posec3d}         & Skeleton                                                    &          23.00        &       83.96   &    21.22     &         85.33         &        15.46          &         91.17         &           15.44       &        90.08    & 18.78 &   87.64   \\

STGCN~\cite{STGCN}                            & Skeleton                                        &         30.60         &         70.32         &        18.33          &           89.97       &     8.76             &       92.44           &          33.39        &        87.22     & 22.77 & 84.99    \\ \bottomrule
\end{tabular}}

\end{table*}

\begin{figure}
    \centering
    \includegraphics[width=0.47\textwidth]{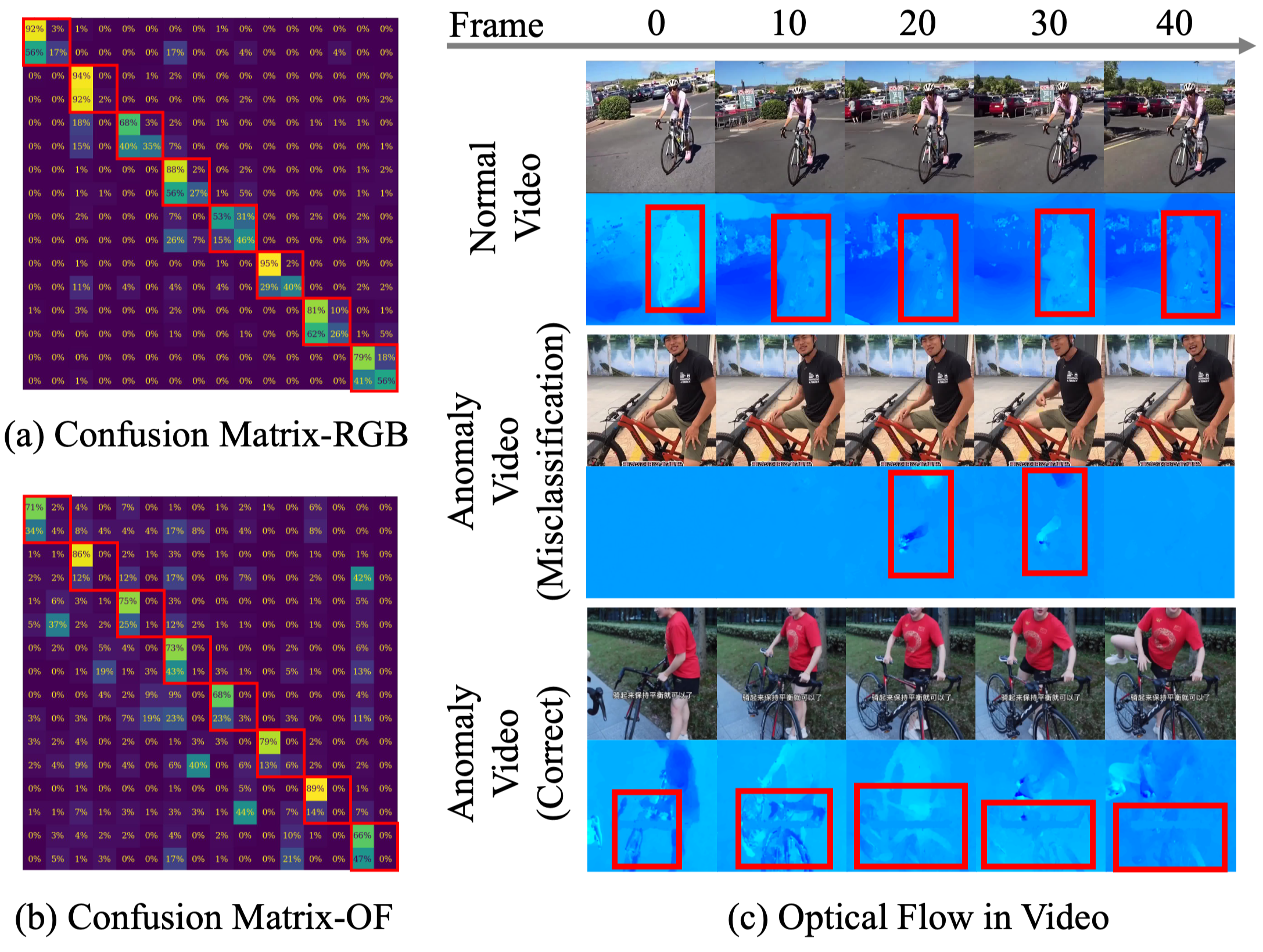}
    \caption{Confusion matrix of the (a) TSN with RGB input and (b) TSN with optical flow field input. Every red block indicates the pairs of videos in the same environment. (c) The optical flow field of the videos in class \textit{``Riding a bicycle"}. The videos from the top to bottom are the normal video, anomaly video that is misclassified in TSN(flow), and anomaly video that is classified correctly in TSN(flow). The numbers above are the frame index of the videos.}
    \label{fig:confuse}
    \vspace{-5mm}
\end{figure}

\noindent\textbf{More Research on FG-BSE Protocol.} To find out the reason for these conflicting observations, we propose the confusion matrix on TSN with different inputs as RGB (Figure~\ref{fig:confuse} (a)) and optical flow fields (Figure~\ref{fig:confuse} (b)). First, for normality videos, RGB shows a higher performance than optical flow fields, which indicates RGB is more effective in extracting environment information.
% However, the optical flow field removes the environment information and leads to unsatisfactory results.
% That is why optical flow failed in environment-aware setting. 
% Incidentally, the utilization of human skeletons results in the exclusion of all environment information, retaining solely action information, contributing to a reduction in performance.
% By the way, human skeletons remove all environment information and only keep action information, which will also degrade performance. 
Furthermore, considering the red blocks which indicate the sub-matrix of behavior pairs with similar environments, we find that a large ratio of anomaly videos is classified into corresponding anomaly videos. This indicates that the inter-environment variance is larger than the intra-environment, and RGB input cannot form a representative feature space. For optical flow fields, most anomaly videos are misclassified to the wrong environments, which presents the incapacity of environment exploration.
% That is why the overall performance is suboptimal in these settings. 

% The comparison result is shown in Table~\ref{setting1_1}. We have the following observations: 1) For all methods with only RGB input, C3D, which is the most simple model, obtains the best result. One reason is that SlowFast and TSM shift the channels in time to force the model to learn temporal information, which is not suitable for our short videos. Another reason is that our data is insufficient for VideoMAE with ViT structure to learn useful information. 2) Methods that introduce the optical flow and human skeletons can obtain a comparable result to the best result in C3D, which indicates that RGB will easily concern with environment information. Human skeleton and optical flow features are enough for the model to recognize the action information in behaviors.  3) With the result in Table~\ref{setting1213}, we found that the methods with only RGB inputs do not perform well. The reason is that RGB can extract more environmental information, which will lead to misclassification in videos with the same environment. Generally, environment-aware settings obtain an inferior result than environment-agnostic settings, which indicates that environment-aware is a hard scenario.

% \begin{table}[t]
% \centering
% \begin{tabular}{@{}c|cccccc@{}}
% \toprule
% \textbf{Frames}     & \textbf{1} & \textbf{2} & \textbf{4} & \textbf{8} & \textbf{16} & \textbf{32} \\ \midrule
% Chronology & 30.22  &   &   &   &  22.66  &    \\
% Random     &   &   &   &   &    &    \\ \bottomrule
% \end{tabular}
% \end{table}

\subsection{Comparsion Result of FG-BSA Protocol}

As shown in Table~\ref{setting21}, for FG-BSA-BC protocol, methods with RGB inputs can easily tackle the issue. 
% However, when comes to modalities like the human skeleton, 
Nevertheless, in the case of the human skeleton, a substantial deterioration in performance is evident. It is obvious the human skeleton fails to capture pivotal environment information. The most intriguing observation is that optical flow fields also show good performance and even outperform some methods with RGB. From the previous analysis, the optical flow field is hard to obtain the environment information.
% , but it can still show promising results in similar action settings.
The following question we need to explore is that \textit{``What kind of environment and action information does the optical flow field extract?"} 
% \begin{figure}[t!]
%     \centering
%     \includegraphics[width=0.4\textwidth]{pics/flow1_cropped_cropped.jpg}
%     \caption{The optical flow field of the videos in class \textit{``Riding a bicycle"}. The videos from the top to bottom are effective video (1st row), environment-effective video that is misclassified in the environment in TSN(flow) (3rd row), and environment-effective video that is classified correctly in the environment in TSN(flow) (5th row). The numbers above are the frame index of the videos.}
%     \label{fig:flow}
%     %\vspace{-3mm}
% \end{figure}
% To fully explore the previous question, we illustrate the optical flow fields of the videos in class \textit{``Riding a bicycle"}. 
To comprehensively investigate the preceding inquiry, we illustrate the optical flow fields extracted from videos belonging to the class labeled \textit{``Riding a bicycle"} in Figure~\ref{fig:confuse} (c).
Considering the normal video in the first row, we found the object (bicycle) that human interacted with is reflected in the optical flow field. Also, we show the anomaly videos that are misclassified (3rd row) and classified correctly (5th row). It indicates that the environment objects will not reflect on the optical flow unless the human interacts with them.
% \textbf{optical flow field can extract the action information and only environment information that the human interacts with}. 
The optical flow field not only captures the action but also encompasses environment associated with human interactions.
This may explain why the optical flow field is one of the desirable inputs in behavior recognition as it removes unmeaningful environment information.
% We also find that the multi-modality input (TSN with the RGB and optical flow field inputs) is also effective and obtains a state-of-the-art result in FG-ESA-BC protocol. One reason is that the multi-modality input provides multi-level features, which provide comprehensive cues for recognition.

\begin{figure}
    \centering
    \includegraphics[width=0.45\textwidth]{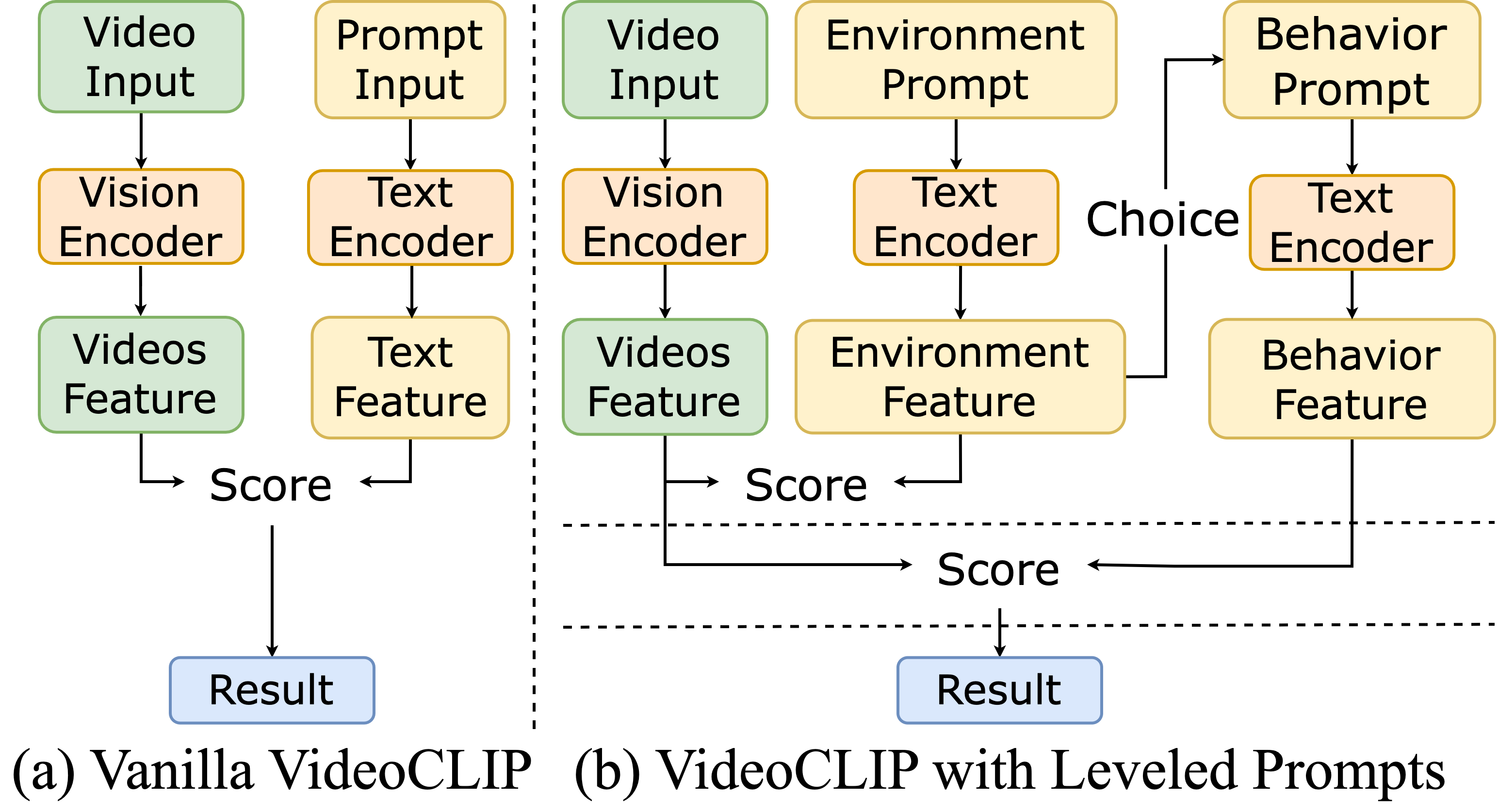}
    \caption{Designs of video-text multi-modality model. (a) vanilla VideoCLIP. (b) VideoCLIP with leveled prompt. First, the model classifies the videos via environment prompts. Then, the video is classified by the prompts of corresponding behavior.}
    \label{fig:clip}
\end{figure}

\subsection{Experiment on Text Modality in Zero-shot}

With the development of CLIP, text modality has started to attract more attention. We utilize VideoCLIP~\cite{videoclip} to explore how the text modality learns effective features.
% Text modality are better at conveying environmental information, compared to describing action information. 
We utilize the text modality to tackle the issue in the FG-BSE-EAW protocol.
% : RGB modality tends to focus on inter-environment variance.

% \subsubsection{Implement Detail}
To evaluate how the text modality contributes to environment information learning, we design two architectures, as shown in Figure~\ref{fig:clip}. The first is vanilla VideoCLIP. The prompt for normality videos is defined as ``Someone is doing \texttt{[BHV]}.", for the anomaly videos, the prompt is defined as ``Someone is not doing \texttt{[BHV]}, but there is \texttt{[ENV]}.".
\texttt{[BHV]} means the behavior, \texttt{[ENV]} means the corresponding environment.
% The encoded video features will measure the score to 16 prompt features.
% and the final classification is determined via maximum softmax score. 
Recognizing the prominent inter-environment variability, we formulate leveled prompts to 
mitigate the excessive concern in environment information during video classification.
% mitigate the excessive concern about environment information when classifying the videos in the same environment. 
At the environment level, we utilize environment prompts as ``There is \texttt{[ENV]}." Then, the problem is changed to a binary classification. We design ``Someone is doing \texttt{[BHV]}." and ``Someone is not doing \texttt{[BHV]}." as behavior prompts. We test on FG-BSE-EAW in a zero-shot manner and the metric is Top-1 accuracy.

% \usepackage{booktabs}
% \begin{table}[t]
% \centering
% \caption{The conclusion of the modality contribution to environment and action learning in behavior recognition. \textcolor{black}{\Checkmark}: Modality can learn this factor. \textcolor{black}{\XSolidBrush}: Modality cannot learn this factor. \textcolor{black}{$\triangle$}: Modality can learn this factor in some condition.}
% \label{summary}
% \scalebox{0.85}{
% \begin{tabular}{@{}c|cc@{}}
% \toprule
% \textbf{Modality}  & \textbf{Environment}                  & \textbf{Action}                            \\ \midrule
% RGB                &           \textcolor{green}{\Checkmark}                            & \makecell{\textcolor{cyan}{$\triangle$}\\When environment \\ information \\ is restricted} \\ \midrule
% Optical Flow Field & \makecell{\textcolor{cyan}{$\triangle$}\\Environment \\ associated with \\
% human interactions} &                   \textcolor{green}{\Checkmark}                         \\ \midrule
% Human Skeleton     &   \textcolor{red}{\XSolidBrush}             &       \textcolor{green}{\Checkmark}             \\ \midrule
% Text               &       \makecell{\textcolor{green}{\Checkmark}\\Still limited}        &         \textcolor{red}{\XSolidBrush}             \\ \bottomrule
% \end{tabular}}

% %%\vspace{-3mm}
% \end{table}

% \subsubsection{Comparison Result}

As shown in Table~\ref{tab:text}, we have the following observations: 1) VideoCLIP with leveled prompts shows a better performance than the vanilla VideoCLIP, which indicates that environment prompts can provide effective environment information. 
% which will ensure the videos classify into the correct environment. 
thereby ensuring accurate classification of videos into their respective environments.
% However, in the context of anomaly videos, the performance degrades. Because environment-effect videos have consistency in the environment but not in actions, particularly given the straightforward nature of our prompt design. Text takes advantage of obtaining environment but fails in actions. 
2) Generally, the performance of text modality is still undesirable. One underlying reason is that VideoCLIP is trained by HowTo100M~\cite{howto100m}, smaller than large-scale image datasets, like LAION-5B~\cite{laion}. The scarcity of high-quality video-text pairs poses a substantial hindrance to the utilization of the text modality for behavior recognition.
% Due to the difficulty in collecting high-quality video-text pairs, it is still limited to utilizing text modality for behavior recognition.

\begin{table}[t]
\caption{The performance of VideoCLIP and VideoCLIP with leveled prompts in the FG-BSE-EAW protocol. We show the accuracy of all videos, normality videos, and anomaly videos.}
\label{tab:text}
\centering
\scalebox{1.0}{
\begin{tabular}{@{}l|ccc@{}}
\toprule
\multirow{2}{*}{\textbf{Method}}        & \multicolumn{3}{c}{\textbf{Top-1 Accuracy(\%)}}                                                      \\ \cmidrule(l){2-4} 
                               & \multicolumn{1}{c}{Total} & \multicolumn{1}{c}{Normality} & \multicolumn{1}{c}{Anomaly} \\ \midrule
VideoCLIP~\cite{videoclip}                      &                   33.63        &            37.71                &                    \textbf{19.94}            \\
VideoCLIP with leveled prompts &           \textbf{34.52}                &         \textbf{39.70}                   &         17.17                       \\ \bottomrule
\end{tabular}}

% %\vspace{-6mm}
\end{table}

\subsection{Summary}\label{summary}

With the previous experiments, we summarize the observations as shown in Figure~\ref{fig:fig1} (b). \textbf{1)} RGB contributes to learning environment information but exhibits limitations in learning action unless the environment is restricted. Notably, it contains the most comprehensive information for behavior recognition. \textbf{2)} The optical flow field demonstrates robustness because it not only learns action information but also the environment associated with human interactions, 
% which ignores meaningless environment information. 
effectively ignoring irrelevant environment information.
\textbf{3)} The human skeleton only learns action in intuitive understanding. \textbf{4)} The text can acquire environment information but still faces the dilemma of degraded performance which is caused by the scarcity of high-quality data.
% compared to previous methods which may be caused by the lack of vast high-quality data.

\section{Conclusion}

This work explores fine-grained behavior recognition from a new standpoint, considering environment and action factors. To address this issue, we construct a new fine-grained behavior dataset BEAR with a well-controlled similar environment and similar action protocols distinguishing it from the previous benchmarks. To address the explorations on how to extract essential features from videos, additionally, we conduct extensive experiments with different modality inputs, including RGB, optical flow, skeleton, and text,
accompanied by analyzing the interplay between modality, environment, and action. With the comprehensive experiments, we provide observations on different modalities for their trends on learning essential factors in video.
These observations collectively offer valuable insights to guide future research endeavors to obtain more informative features from the videos.

\section*{Acknowledgment}
This work is supported by YuCaiKe[2023] Project Number: 14105167-2023, the Fundamental Research Funds for the Central Universities (YG2023QNA35) and National Natural Science Foundation of China (No. 62472282).

\bibliographystyle{IEEEbib}
\bibliography{icme2025references}

\end{document}